\documentclass[conference]{IEEEtran}
\IEEEoverridecommandlockouts
\usepackage{cite}
\usepackage{amsmath,amssymb,amsfonts}
\usepackage{algorithmic}
\usepackage{graphicx}
\usepackage{textcomp}
\usepackage{xcolor}
\usepackage{hyperref}
\usepackage{subcaption}
\def\BibTeX{{\rm B\kern-.05em{\sc i\kern-.025em b}\kern-.08em
    T\kern-.1667em\lower.7ex\hbox{E}\kern-.125emX}}
\begin{document}

\title{Verifiable Split Learning via zk-SNARKs
\thanks{This work was supported by the Spanish Government under these research projects funded by MCIN/AEI/10.13039/501100011033: DISCOVERY PID2023-148716OB-C31; the grant "TRUFFLES: TRUsted Framework for Federated LEarning Systems", within the strategic cybersecurity projects (INCIBE, Spain), funded by the Recovery, Transformation and Resilience Plan (European Union, Next Generation)"; and the Galician Regional Government under project ED431B 2024/41 (GPC). Finally, we would like to thank the Centro de Supercomputación de Galicia (CESGA) for their support and the resources for this research.}
}

\author{\IEEEauthorblockN{Rana Alaa \IEEEauthorrefmark{1}, Darío González Ferreiro\IEEEauthorrefmark{1}, Carlos Beis-Penedo\IEEEauthorrefmark{1}, \\ M. Fernández-Veiga\IEEEauthorrefmark{1}, Rebeca P. Díaz-Redondo\IEEEauthorrefmark{1}, Ana Fernández-Vilas\IEEEauthorrefmark{1}}
\IEEEauthorblockA{\IEEEauthorrefmark{1}Information \& Computing Lab,
atlanTTic, Universidade de Vigo}}


\maketitle

\begin{abstract}
Split learning is an approach to collaborative learning in which a deep neural 
network is divided into two parts: client-side and server-side at a cut layer. 
The client side executes its model using its raw input data and sends the intermediate 
activation to the server side.  This configuration architecture is very useful for 
enabling collaborative training when data or resources are separated between devices.
However, split learning lacks the ability to verify the correctness and honesty of 
the computations that are performed and exchanged between the parties. To this purpose, 
this paper proposes a verifiable split learning framework that integrates a zk-SNARK 
proof to ensure correctness and verifiability. The zk-SNARK proof and verification 
are generated for both sides in forward propagation and backward propagation on the 
server side, guaranteeing verifiability on both sides. The verifiable split learning
architecture is compared to a blockchain-enabled system for the same deep learning 
network, one that records updates but without generating the zero-knowledge proof.  
From the comparison, it can be deduced that applying the zk-SNARK test achieves 
verifiability and correctness, while blockchains are lightweight but unverifiable.
\end{abstract}


\section{Introduction}
Split learning (SL) is a collaborative training model that splits a neural network at
an agreed layer called cut layer keeping the first layers on the client device and 
the deeper layers on a remote server~\cite{khan2023love}. In this way, the raw data 
remains local and only the "smashed data" intermediate activations are transmitted to 
the server. Currently, split-learning processes do not have any mechanism to confirm 
that each part correctly performs the computation assigned to it. On the client side, 
the server must accept incoming activations without proof that they have been 
generated by the agreed model on real data. On the server side, the client must 
trust that the step forward, loss calculation and gradient updates are executed
correctly and returned unchanged~\cite{tajabadi2024privacy}. A malicious or 
faulty client could fabricate activations, while a malicious server could 
return incorrect gradients or skip updates all together. Since neither party can
independently verify the work of the other, the integrity of the joint training
process remains unproven. This mutual uncertainty opens the door to silent 
failures and adverse behaviour ---fabricated activations, poisoned gradients or
missed updates--- that can corrupt training or leak information.  Current 
split-learning pipelines preserve the locality of data, but cannot prove the 
integrity of distributed computations and its verifiability. So, this 
restriction prevents the technology from being deployed in high-risk domains, 
e.g. in healthcare, finance, or regulated IoT, where participants must prove 
both privacy protection and strict correctness of each training
step~\cite{pasquini2021unleashing,pham2024enhancing}. To tackle this restriction,
cryptographic approaches can be deployed in SL to  ensure correctness  and
verifiability without revealing sensitive data at the same time, like zk-SNARK.

Zero-knowledge, succinct, non interactive arguments (zk-SNARKs) are a type of 
cryptographic proof that provide a way for a prover to convince a verifier that 
some computation has been performed correctly without revealing any private data.
zk-SNARK can be instantiated by several methods~\cite{ayoub2025verifbfl}, capable
of providing  constant test sizes and fast verification. Therefore, when 
verifiability is the priority,  zk-SNARKs offer the lighter, scalable option.
As stated, one limitation of split learning is that it is not verifiable: clients 
cannot be sure that the server has performed the calculations correctly, and the 
server has the same uncertainty with regard to the client, which entails additional
costs and weakness trust. Zero-knowledge proofs, offer a concise way to certify 
correctness at reasonable computational cost. 

In this paper, to address the trust gap in split learning, we add cryptographic
verifiability: each party attaches a zk-SNARK proof to its part of the computation.
This is further compared with a blockchain-enhanced SL system, as blockchains can
register updates immutably but do not guarantee that the computations are 
performed correctly.
Our main contributions are the following. First, we integrate a two-sided zk-SNARK scheme into a decentralized split 
learning architecture, both for the forward and backward exchange of parameters 
between the two sides of the split neural network. We also leverage a simple arithmetic circuit to provide integrity throughout the SL 
learning cycle. Finally, We measure the performance and computational cost of our verifiable SL and 
compare the results with a SL system without embedded integrity protection but that 
relies on an external blockchain to support trust and immutability. 


\section{Background}
\label{sec:background}

The following background reviews the essential concepts underlying this study, 
namely split learning, zero-knowledge proofs, and blockchain

\subsection{Split learning}

In SL~\cite{khan2023love}, deep neural networks (DNNs) are split into several 
portions, and these separate sub-networks exist  and run on various computing 
devices. Splitting a DNN into two portions is the most straightforward SL 
approach. The first sub-network is deployed on the client side and the second 
sub-network is deployed on a host server. The last layer of the network part on 
the client side is called the cut layer. At the beginning, the first client and 
the host server randomly initialize weights for the neural network  part they 
possess. In forward propagation, the clients compute sequentially with their 
own  data up to the cut layer, and then send the activations, called smashed data, 
to the server. Since the clients' private data is not transmitted to another server,
individual privacy is guaranteed. After receiving the activations from the client, 
the host server continues training the part of the DNN in its possession. When the 
host server completes the forward propagation phase, it calculates the gradients 
and conducts the backward propagation. The gradients in the first layer of the 
server-side portion of the model are sent back to the client to complete the
backpropagation phase in the rest of the neural network. After the completion 
of forward and backward propagation on the local data, the client updates its 
local model, which is sent to the next client, for the next round of training as 
shown in the following part while explaining the formal split learning 
definition~\cite{tran2022privacy}. 

Thus, split learning decomposes a global DNN as
\begin{equation*}
  f(\mathbf{x};\mathbf{w})
  = f_s\bigl(f_c(\mathbf{x};\mathbf{w}_c);\mathbf{w}_s\bigr),
  \quad \mathbf{w}=[\mathbf{w}_c;\mathbf{w}_s]
  \label{eq:model_partition}
\end{equation*}
where $f_c:\mathcal{X}\to\mathcal{Z}$ runs on the client and
$f_s:\mathcal{Z}\to\mathcal{Y}$ runs on the server~\cite{Vepakomma2019,Gupta2018}.
The learning algorithm in SL then proceeds as follows:
\begin{enumerate}
\item \textbf {Forward Pass}. For each $(\mathbf{x}_i,y_i)$ in batch $B$, 
$\mathbf{z}_i = f_c(\mathbf{x}_i;\mathbf{w}_c)$, and $\hat{y}_i = 
f_s(\mathbf{z}_i;\mathbf{w}_s)$. Clients transmit only $\mathbf{z}_i$ 
(“smashed data”) to the server~\cite{Vepakomma2019}.

\item \textbf {Loss and Gradients}. Define per‐sample loss $L_i = 
\ell\bigl(\hat{y}_i,y_i\bigr)$. The server computes $\mathbf{g}_{w_s}^{(i)} = 
\nabla_{\mathbf{w}_s}L_i$, $\mathbf{g}_{z}^{(i)} = \nabla_{\mathbf{z}_i}L_i$,
and returns \(\mathbf{g}_{z}^{(i)}\) to the client~\cite{Vepakomma2019}.

\item \textbf {Backpropagation and Updates}. Clients apply the chain rule:
\begin{equation*}
  \mathbf{g}_{w_c}^{(i)}
  = \bigl(\nabla_{\mathbf{w}_c}f_c(\mathbf{x}_i;\mathbf{w}_c)\bigr)^{\!\top}
    \mathbf{g}_{z}^{(i)}
  \label{eq:client_grad}
\end{equation*}
and updates the parameters over batch $B$:
\begin{align*}
  \mathbf{w}_s &\leftarrow \mathbf{w}_s
    - \eta\,\frac{1}{|B|}\sum_{i\in B}\mathbf{g}_{w_s}^{(i)}, 
  \label{eq:update_ws}\\
  \mathbf{w}_c &\leftarrow \mathbf{w}_c
    - \eta\,\frac{1}{|B|}\sum_{i\in B}\mathbf{g}_{w_c}^{(i)},
\end{align*}
with learning rate \(\eta\) \cite{Vepakomma2019}.
\end{enumerate}

\subsection{Zero-knowledge proofs}

Zero Knowledge Succinct Non-interactive Arguments of Knowledge (zk-SNARK) are a type of zero-knowledge proofs characterised by their small proof size and fast verification times~\cite{sheybani2025zero,nitulescu2020zk,ballesteros2024enhancing,chen2022review}. These schemes are defined by a relation $R$ between a witness $w$ and a statement $x$ such that if the relation between $w$ and $x$ holds, then $(x, w) \in R$. A security parameter $\lambda$ can be derived from the description of $R$. The prover needs to reveal only $x$ to the verifier, and no information about $w$ is revealed in the proof.

A zk-SNARK scheme is composed by the following operations:

\begin{itemize}
    \item $(crs=(pk, vk), td) \gets \mathsf{Setup}(1^\lambda, R)$: Takes a relation $R$ and a security parameter $\lambda$ and outputs the Common Reference Key $crs$, which is divided into the Proving Key $pk$ and the Verification Key $vk$. Also, it outputs a trapdoor $td$.
    \item $\pi\gets \mathsf{Prove}(pk, x, w)$: From $pk$, a witness $w$ and a statement $x$, a proof $\pi$ is generated. 
    \item $Accept/Reject \gets \mathsf{Verify}(vk, x, \pi)$: From $vk$, a proof $\pi$ and the corresponding statement $x$, outputs $Accept$ or $Reject$. In the context of this work, a \textit{valid proof} refers to any $(x_v, \pi_v)$ such that $\mathsf{Verify}(vk, x_v, \pi_v) = Accept$.
    \item $\pi \gets \mathsf{Sim}(crs, td, x)$: Employs the trapdoor $td$ to simulate a proof $\pi$ for statement $x$.
\end{itemize}

A zk-SNARK scheme possesses the following security properties:

\textbf{Completeness}. The probability 
\[
\operatorname{Pr}
\left[
\begin{array}{c}
((pk,vk), td) \gets \mathsf{Gen}(R) \\
\pi \gets \mathsf{Prove}(pk, x, w) \\
(x, w) \in R
\\\hline
\mathsf{Verify}(vk, x, \pi) = Reject
\end{array}
\right]
\]
is negligible. Intuitively, this means that an honest prover is able to convince a verifier that $(x, w) \in R$.

\textbf{Knowledge Soundness}. For every efficient adversary $A$, there exists an efficient extractor $Ext_A$ with access to the internal state of $A$ such that the probability 
\[
\operatorname{Pr}
\left[
\begin{array}{c}
((pk,vk), td, aux) \gets \mathsf{Gen}(R) \\
(x, \pi) \gets \mathsf{A}(R, aux, (pk, vk)) \\
w \gets Ext_A(R, aux, (pk, vk))
\\\hline
(x, w) \notin R \\
\land \mathsf{Verify}(vk, x, \pi) = Accept
\end{array}
\right]
\]
    is negligible, where $aux$ is an auxiliary input produced by $\mathsf{Gen}$. Intuitively, this means that dishonest provers could not generate a valid proof if they do not know $w$.

\textbf{Zero-Knowledge}. For every adversary A acting as a black box and $(x, w) \in R$, there exists a simulator $\mathsf{Sim}((pk, vk), aux, x)$ such that the following equality holds:
\[
\Pr\left[
\begin{array}{c}
(pk, vk), td, aux \gets \mathsf{Gen}(R) \\
\pi \gets \mathsf{Prove}(pk, x, w) \\
\hline
\mathsf{A}((pk, vk), aux, x, \pi) = 1
\end{array}
\right]
\]
$\approx$
\[
\operatorname{Pr}
\left[
\begin{array}{c}
((pk,vk), td, aux) \gets \mathsf{Gen}(R) \\
\pi \gets \mathsf{Sim}((pk, vk), td, x) \\
\hline
\mathsf{A}((pk, vk), aux, x, \pi) = 1
\end{array}
\right]
\]
where $aux$ is an auxiliary input produced by $\mathsf{Gen}$. Intuitively, this means that an attacker cannot find out anything about a witness $w$ from a proof $\pi$, a statement $x$ and a key pair $(pk, vk)$.

\subsection{Blockchain}

Blockchain is a cutting-edge technology that reconfigures transactions at its core 
and interacts with several entities, such as institutions and governments, and 
ensures authenticity processes. Blockchain technology is 
known for implementing a decentralised log  to maintain an unalterable record of
transactions~\cite{liu2024enhancing}. Thus, the blockchain provides a proof of 
existence for opaque data blocks, including timestamps and associated metadata.
Transactions stored in the blockchain are electronically signed and archived in 
a distributed filesystem. In this way, the recorded history of all transactions
can be recorded in a secure manner.  Blockchains can be public or 
permissionless blockchain, private or permissioned blockchain and consortium 
blockchain. In a public blockchain, there is no leading authority and no entity 
has higher authority than others in the distributed ledger. Participants can enter 
and leave at any time if they want to. Likewise, any participant can check the
transaction as it is public. A private blockchain adopts a 
centralised structure where a unique entity has all the control to check 
transactions and take decisions. The private blockchain is more efficient, less 
complex to implement, requires reduced energy resources and is faster when 
compared with the public blockchain~\cite{qammar2023securing}.
 
 \section{Related Work}
 \label{sec:related Work}

 
In the field of federated learning, the  
articles~\cite{ li2023martfl,ahmadi2024zkfdl,wang2024zkfl} illustrate that 
public verification of aggregation based on zk-SNARKs is practical: the proofs 
remain small and verifications are fast, but the costs are concentrated on the 
prover and communication costs may increase.

for split learning, recent works as~\cite{higgins2025towards} 
and~\cite{pham2024enhancing} demonstrate the possibility of introducing 
improvements in privacy and bandwidth ---by protecting features such as Class
Activation Mapping (CAM) and Adaptive Differential Privacy (ADP). CAM 
can identify and challenge adverse inference and reconstruction adversarial 
attacks in split learning by detecting the sensitive or vulnerable 
features~\cite{higgins2025towards}, whereas ADP dynamically sets the 
differential privacy noise level based both on the training process and the 
properties of the data for balancing accuracy and privacy~\cite{pham2024enhancing}.  

However, these approaches focus on preserving confidentiality and 
lack possibility of verifying the correctness of distributed computations.
As~\cite{pasquini2021unleashing} illustrates in the context of split-learning 
security analysis, the leakage of important information from intermediate 
features through reconstruction attacks and attribute inference attacks is 
possible, demonstrating that privacy mechanisms alone are not sufficient to 
guarantee auditability or correctness.

In general, federated learning using zk-SNARKs is able to provide verification 
but  suffers from high computational costs, while split learning  is more 
efficient, but also requires more strong verification mechanisms. 
Furthermore, privacy defense mechanisms do not support the validity of 
calculations.  Hence, the work presented in this paper is to address these 
concerns by presenting  a novel approach that integrates SL with Groth16 
zk-SNARK proofs to prove the validity of forward and backward propagation 
during training. This approach differs from other research because it provides 
the possibility of verifying correctness while maintaining privacy and 
transparency. In addition, it allows an external  verifier to validate the
correctness of the results without having to access the raw data.

\begin{figure}[t]        
 \centering
  \includegraphics[width=0.9\columnwidth]{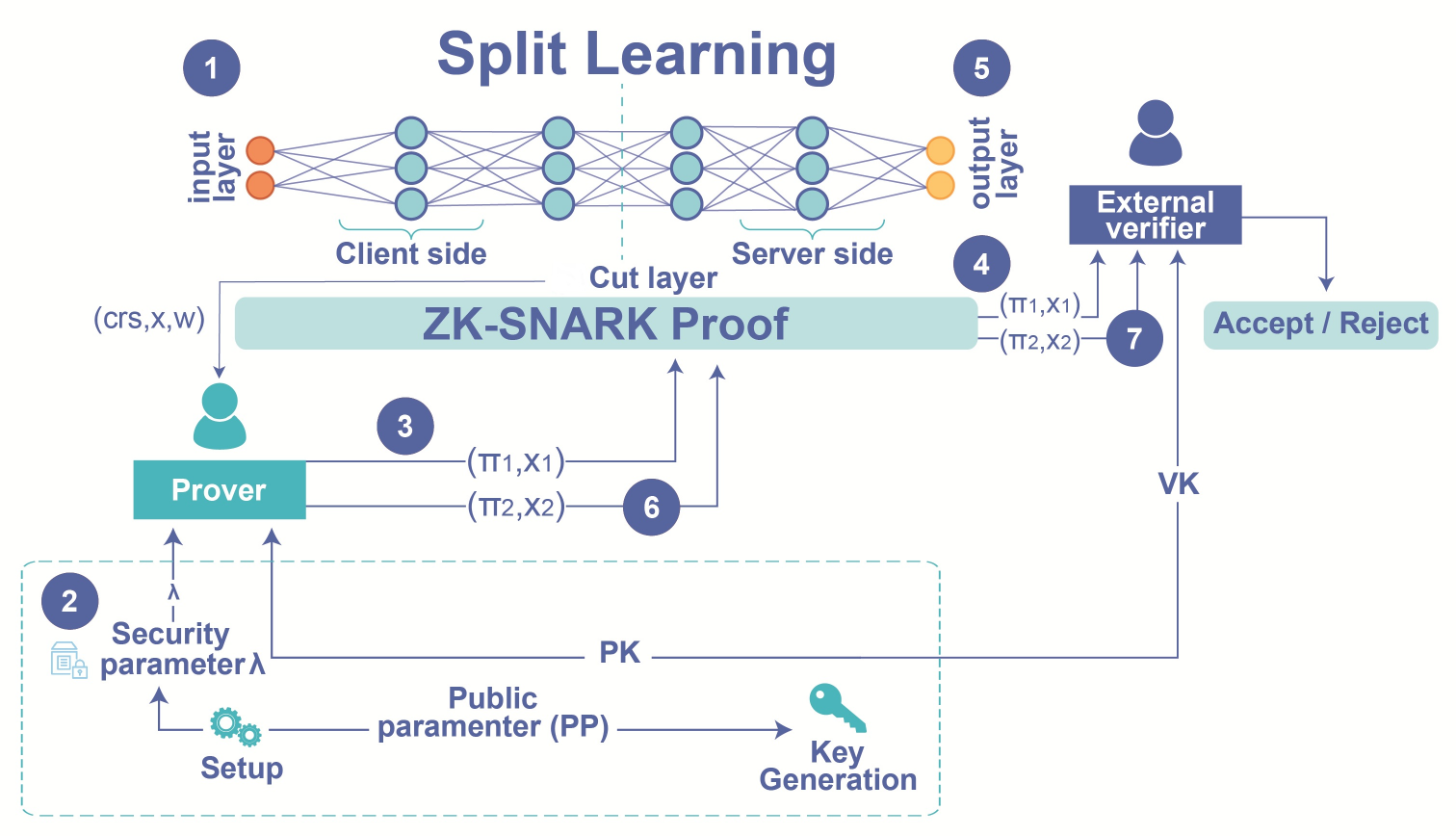}
  \caption{Proposed split learning with zk-SNARK system architecture}
 \label{fig:fig1}
\end{figure}

\section{System Design}
\label{sec:system architecture} 

\subsection{System Model}

The entities that take part in our proposed architecture are represented in 
Figure~\ref{fig:fig1}. Split learning of a given neural network (NN) model
is divided at a predefined  ‘cut layer’, and the \textbf{worker nodes} participating 
in the system execute only one of the two sides of the NN. We refer to the part 
of the split that receives the input data as client-side, and to the other part 
as server-side. Thus, the client side takes the raw data and executes its part 
of the model, while the server side hosts the deeper layers. For the computation
and verification of the proofs, and for the setup of the initial security parameters,
an external \textbf{Prover entity} (PE) is  responsible for generating 
zero-knowledge proofs for the vectors of parameters exchanged between a client and
a server worker nodes. The task of verifying these proofs is separately run by 
the \textbf{Verifying Entity} (VE), who receives requests and proofs from the 
worker nodes and returns a result of acceptance or rejection of the proof. Both 
the PE and the VE are assumed to be trusted parties, and moreover the communication
between a worker node and the PE takes place on a secure private channel. 
Note, however, this assumption is not critical, since our description presents 
only the logical architecture of the system: provided the  worker nodes have 
enough computational power, the functions of the PE (excluding the setup) could 
be embedded into them. 

We assume a decentralized split learning system with an arbitrary number of nodes, 
possibly with several clients and servers. The communication infrastructure between 
clients and servers is provided by a network operator, and the strategy for 
executing the aggregation of parameters at the servers who receive messages 
from several clients will be described later.

\subsection{Workflow}
The workflow for execution of training and inference tasks on this system proceeds 
as follows.  During the training phase  (step 1), a client 
side processes its forward pass layers on the private input data, creating the
smashed data (activation vector). The next step (step 2 in Fig.~\ref{fig:fig1}) 
begins by building the setup phase  and generating the keys for both the PE (PK) 
and the VE (VK) to certify the correctness of the calculation with the updated
weights that the client generated. Subsequently, in step 3, a Groth16 zk-SNARK 
proof is generated for these weights using the private channel between the worker 
and the PE, and the proof is sent along with the data to a server-side worker node.
This worker node verifies the proof that the data are honest using the VE, and
accepts the incoming message if the verification succeeds (steps 4 and 5).
Upon success, the server side execute the remaining layers of the forward pass,
calculates the loss, and computes its corresponding backward pass stages to o
obtain the gradients. This is followed by step 6, where a new Groth16 zk-SNARK 
proof is generated through the PE, such that the clients receiving this update 
can eventually verify with the VE the integrity of this backward vector of parameters 
(step 7). Invalid or missing proofs are discarded, and the client is 
excluded from this round if the external verifier rejects it. Finally, the verified 
and updated gradients are sent back through the cut layer encrypted to allow updating
of the parameters. By embedding the  zk-SNARK proof and verification on both 
sides of the SL, each forward and backward calculation is guaranteed to 
be honest, while keeping the users' raw data and full model parameters private,
something that the split learning alone cannot guarantee. Further more, by applying
that integrity is guaranteed as the poisoned updates slipped through and also we 
will ensure privacy preserved by revealing nothing about data or model weights in 
the proof.

\subsection{Threat Model}

The main property to guarantee in the proposed scheme is the verifiability on
all the computations done by the worker nodes (clients o servers) in this 
decentralizaed SL system.  Verifiability is defined here as the ability of a 
training node or aggregator (worker nodes) )to prove to other worker nodes
participating in the SL protocol that their part of computation in the split 
NN has been executed honestly on the raw input data, without leaking the private 
data upon used for training or inference.  zk-SNARKs provide this verifiability 
test in out case.

This ability of verifying the computation protects the systems against gradient 
injection or features injection attacks, by which a malicious participant could 
inject adversarial smashed data (either in the forward or in the backward passes)
into the training. It also protects against poisoning attacks where malicious 
clients manipulate the input data to bias the model: this can be simply achieved 
by pre-distributing among the honest participant a small subset of witnesses, i.e.,
common and validated input samples to be used during the training, unknown to the 
attackers. Zerp-knowledge proofs are reciprocal, in that they also protect against
compromised server-side worker nodes who are manipulating the gradients o purpose.

\section{zk-SNARK Proof Construction} 
\label{sec:implementation} 


\paragraph{Quantization}
As ZKPs work with integers, and most of the DNN values are floating-point values, 
it is necessary to quantize them. Suppose a floating point value $x$ so that 
$x \in [a, b]$. It is possible to represent it as an integer $q$ so that  
$q \in [a_q, b_q]$ being  $a_q,\,b_q$ integer  values, just by applying the 
following formula $q \;=\; \Bigl\lfloor \tfrac{x}{s} \Bigr\rfloor + z$,
where $x$ the zero-value and s the scaling factor. The zero-value is the result of
quantizing a zero, and the scaling factor is a value that, noting that quantization 
is a type of scaling, changes the scale. On the other side, we have the dequantization
formula $x   \;=\; s\,\bigl(q - z\bigr)$.
As this quantization may result in overflow, it is needed to increase the range of  
$x$ by a small $\varepsilon$  so that $x \in (\, a - \varepsilon,\; b + \varepsilon\,$.
In order to calculate the values of $z$ and $s$ it is necessary to solve a 
two-equation system, dequantizing $a_q$ and $b_q$, namely $a - \varepsilon = s.\,(a_q - z)$, $b + \varepsilon = s\,(b_q - z)$.

\paragraph{Arithmetic circuit}
Groth16 is a zk-SNARK library that uses an arithmetic circuit to conduct the ZKP, 
which is an equation with some input values and some private values. In this case, 
the arithmetical expression is as follows $W' = W + K\cdot U$
where $W$ is the globalized model in the previous epoch, $W^\prime$ is the 
globalized model in the actual epoch, $K$ is a vector of the weights of each node, 
and $U$ is a matrix containing the model of each node. $W^\prime$ and $W$ are 
$1 \times m$ matrices, while $K$ is a $1 \times n$ matrix  and $U$ is a $n \times m$
matrix,   being $n$ the number of nodes and $m$  the number of variables in the model.
The public inputs are $W^\prime$ , $W$ and $K$, while $U$ remains private. As we 
are implementing Split Learning and the proof is conducted on the Split Layer, $m$ 
is equal to the number of gradients in the Split Layer and $n$ is equal to one, as 
each node only has the information of its own model. This could result in a privacy
issue, so each node generates a pair of public and private keys in order to cipher 
the communication between nodes. This arithmetic circuit is divided into two 
sub-circuits: the aggregation circuit and the update circuit. The aggregation 
circuit is as follows $U' = K \cdot U$.
As ZKPs only work with integer values, these are the quantized versions of 
those matrices. But they are quantized using different parameters, so the scales 
are different. In order to make the numbers match, the dequantization formula is 
needed
\begin{equation*}
    s_{U'} \cdot \bigl(U'_{i,j} - z_{U'}\bigr)
\;=\;
s_{K} \cdot \bigl(K_{i,j} - z_{K}\bigr)\,
s_{U} \cdot \bigl(U_{i,j} - z_{U}\bigr)
\end{equation*}
where $Z_{U'}$, $S_{U'}$, $z_{K}$, $s_{K}$, $z_{U}$ are the zero-value and scaling 
factors of $U^\prime$, $K$ and $(U$ respectively. Developing this expression, in 
order to avoid working with non-integer and negative values, the outcome is as follows
\begin{align*}
    2^{\eta}\,U'_{i,j}
  = R^{a}
  + 2^{\eta} \cdot z'_{U}
  + \Bigl\lfloor
      2^{\eta} \cdot 
      \frac{s_{K}\,s_{U}}{s_{U'}}\,
      \cdot \bigl(M_{1}+M_{4}-M_{2}-M_{3}\bigr)
    \Bigr\rfloor   
\end{align*}
where $M_{1} = \sum_{k=1}^{n} K_{i,k}\,\cdot U_{k,j}$, $M_{2} = z_{U}\ \cdot \sum_{k=1}^{n} K_{i,k}$, $M_{3} = z_{K}\ \cdot \sum_{k=1}^{n} U_{k,j}$, and $M_{4} = n\ \cdot z_{K}\,\cdot z_{U}$,
being \(R^{a}\)  a leftover matrix, which compensates quantization errors, and 
$\eta$ being a big integer (22 or bigger). Following into the update circuit we have
$W' = W + U'$.

As before, applying the dequantization formula leads to
\begin{equation*}
    s_{W'} \cdot \bigl(W'_{i,j} - z_{W'}\bigr)
\;=\;
s_{W} \cdot \bigl(W_{i,j} - z_{W}\bigr)
\;+\;
s_{U'} \cdot \bigl(U'_{i,j} - z_{U'}\bigr)
\end{equation*}
and, by developing this expression, we get
\begin{equation*}
    2^{\eta} \cdot W'_{i,j}
  \;=\;
  R^{u}
  \;+\;
  2^{\eta} \cdot z'_{w}
  \;+\;
  \Bigl\lfloor
     2^{\eta} \cdot 
     \bigl(N_{1}+N_{3}-N_{2}-N_{4}\bigr)
  \Bigr\rfloor
\end{equation*}
where $N_{1} = \frac{s_{w}}{s_{w'}} \cdot W_{i,j}$, $N_{2} = \frac{s_{w}}{s_{w'}} \cdot z_{w}$, $N_{3} = \frac{s_{u'}}{s_{w'}} \cdot U'_{i,j'}$, and $N_{4} = \frac{s_{u}}{s_{w'}} \,\cdot\, z _{U'}$ with $R^{u}$ being another leftover matrix.

\begin{figure}[!t]
  \centering
  \includegraphics[width=0.9\columnwidth]{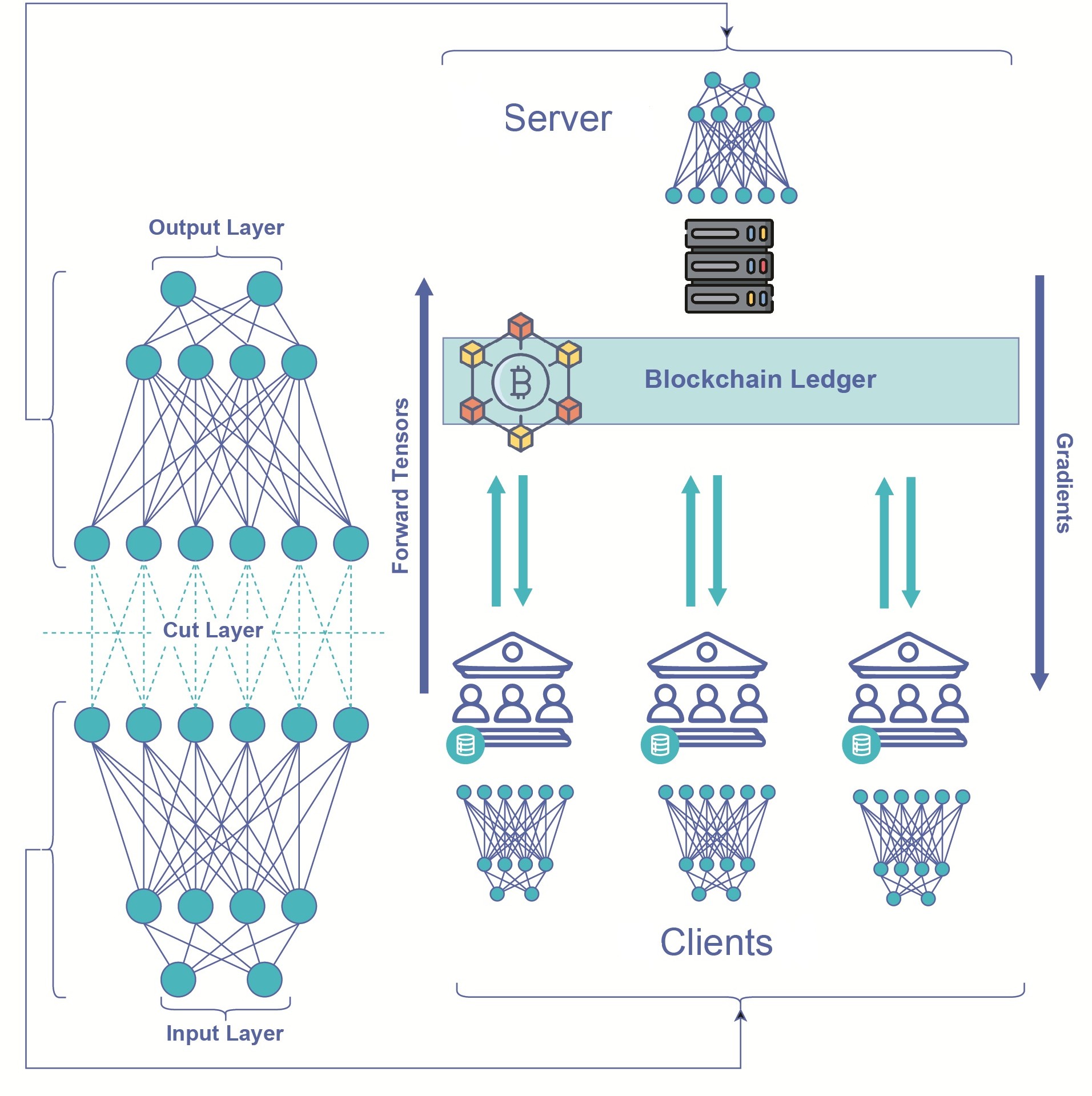}
  \caption{Blockchain overview architecture}
  \label{fig:blockchain}
\end{figure}

Figure~\ref{fig:blockchain} illustrates the second architecture that are used to compare our experimental tests in which the  split learning is integrated with the blockchain. In this architecture, the split
learning is divided into two sides by a cut layer: the client side and the server 
side. The client side processes the raw input data to the cut layer by generating 
intermediate outputs known as forward activations. The client-side output is sent 
to the server side. This transfer is transmitted through the blockchain layer. 
The server receives the client-side output and completes the forward computation 
by calculating the loss and gradients through backpropagation. Afterward, the 
gradients are sent back to the client to update its local model. These 
communications are all recorded in the blockchain ledger.

\section{Experimental results}
\label{sec:experimental results} 

\begin{figure*}[t]
  \centering
  \subcaptionbox{Batch time \label{fig:batchtime}}{%
    \includegraphics[width=0.45\textwidth]{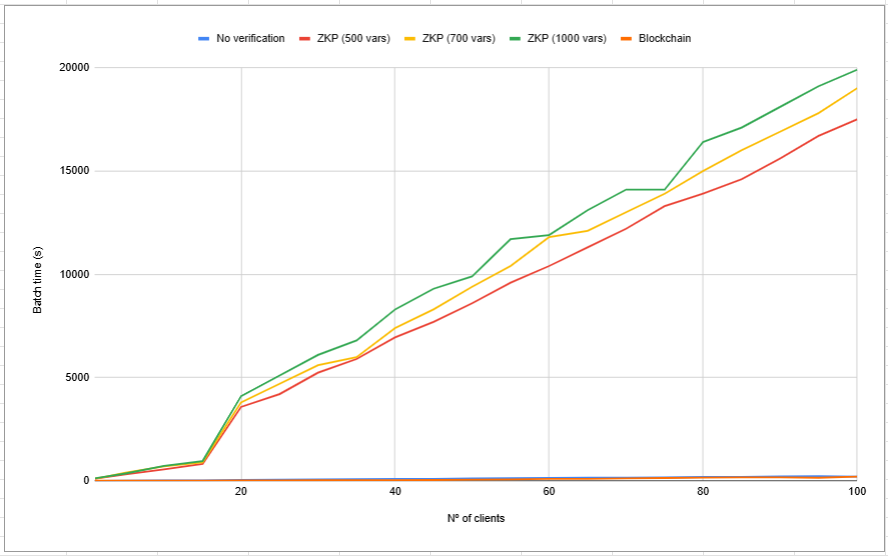}}
  \subcaptionbox{Real epoch estimation time \label{fig:realepochestimationtime}}{
    \includegraphics[width=0.45\textwidth]{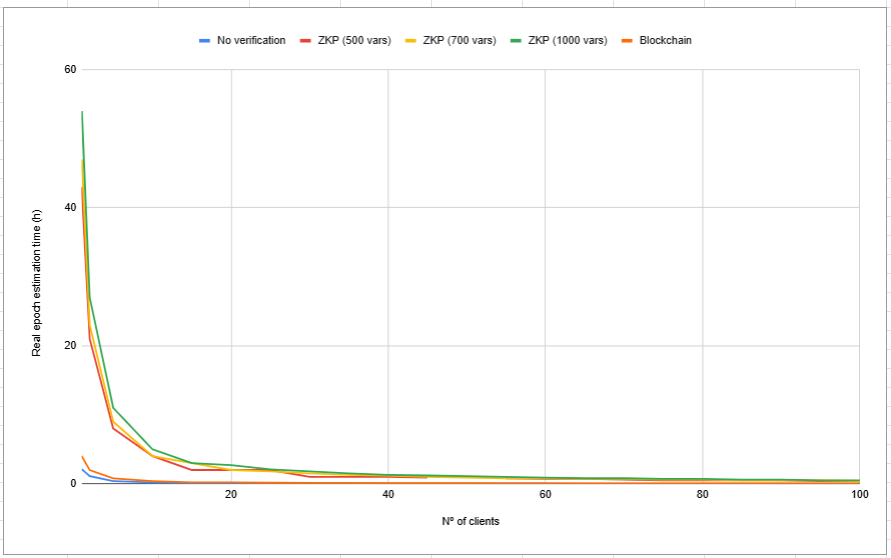}}
  \subcaptionbox{Epoch estimation time \label{fig:epochestimationtime}}{
    \includegraphics[width=0.45\textwidth]{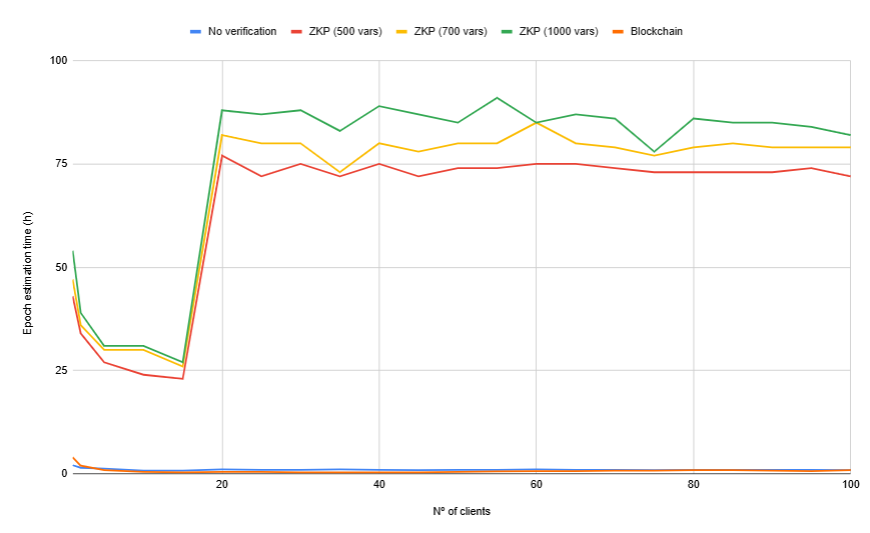}}
  \subcaptionbox{Proof and verification time \label{fig:proofandverificationtime}}{
    \includegraphics[width=0.45\textwidth]{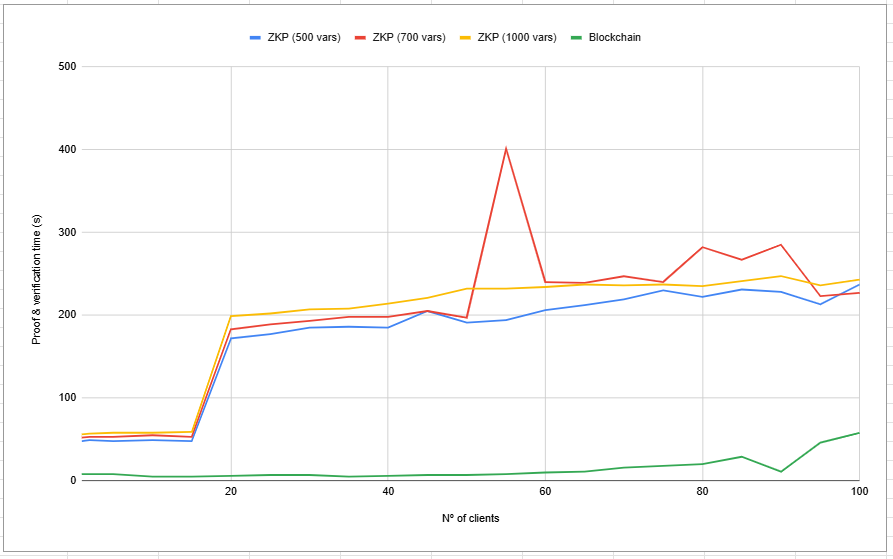}}
    \caption{Performance results and comparison to blockchain support.}
\end{figure*}

This study introduces a simultaneous, privacy-preserving split-learning pipeline
augmented with zero-knowledge proofs (ZKPs) from Groth16. Multiple logical clients 
are instanced as threads sharing a single GPU/CPU, each involving the shallow stem 
of a ResNet-18, while a simulated centralized server maintains the deeper 
convolutional blocks and the classifier overhead. The clients train on disjoint
independent and identically distributed (IID) partitions of 
CIFAR-10~\cite{krizhevsky2009learning}, which are widely and well-known benchmark
dataset. It is composed of 60,000 $32 \times 32$ 
color images across 10 mutually exclusive classes and offers a diverse range of 
images specially suitable for assessing the performance of image classification. 
The images belong to $10$  classes (6,000 images per class). These partitions  
are generated by shuffling the  50,000 training indices with  a fixed set of 
seeds and splitting them into equal-sized Data Loaders (batch size is 32). 


The experimental tests were mainly focused on ZKP and then on a comparison between 
ZKP and blockchain. The experimental tests were performed on two different computing
environments due to technical limitations. The ZKP experiments was performed on
workstation with 32GB RAM for up to 15 clients, while the experiment with more than 
15 clients was performed on FinisTerrae III supercomputer, hosted at CESGA (Centro 
de Supercomputación de Galicia). Furthermore, the blockchain
comparison was also performed between these two different environments. During the
blockchain tests, the blockchain tests were performed on workstation with 32GB RAM 
DDR5 RAM. Therefore, variations in the hardware configuration and the computing 
environment had some impact on the variation of the results. Since the purpose of 
the analysis is meant to check scalability, the analysis of the results will
concentrate on the relative time rather than the absolute time of the computations. 
The analysis of the results demonstrates the changes in performance with the number 
of clients and the complexity of the tests. These experimental tests were performed 
to analyse the batch time, actual epoch estimation time, epoch estimation time, 
test generation and verification time between different complexities (500, 700, 1000)
for various numbers of clients from 1 to 1000.


\subsection{Batch time analysis}

\subsubsection{Scalability}

Figure~\ref{fig:batchtime} present batch times  for various ZK-SNARK setups.  According 
to the results, as the number of clients increases, the batch execution 
time also increases incrementally for all zk-SNARK setups. A clear increase is observed 
for tests of higher complexity,  as circuits with 1,000 variables are showing 
consistently higher computation times when compared to circuits with $500$ or $700$
variables. It is due to the increasing number of clients, implying that more  proofs 
need to be generated at the same time, and multiple proofs require significant  
computation time.
This observation reflects the computationally intensive test generation process as
both the number of clients and circuit complexity are increased. 
Thus, it suggests a limitation on scalability: while zk-SNARKs perform high 
verifiability, their computational cost rises rapidly with system size and 
proof complexity.


\subsubsection{Blockchain comparison}. Figure~\ref{fig:batchtime}  states that 
the batch execution time of the blockchain remains nearly the same compared to 
zk-SNARK setups, which constantly increase as the number of clients increases. 
The reason for this difference is because the blockchain only records updates 
without generating proofs, as compared to zk-SNARK setups, where each client needs 
to generate a proof for each batch. Although the increase is more notable in the 
more complex circuits, since proofs are more complex to generate.


\subsection{Epoch estimation time analysis}

\subsubsection{Scalability}

Figure~\ref{fig:epochestimationtime} presents the epoch estimation time in 
different zk-SNARK configurations. The results indicate that as the number of 
clients increases, the epoch execution time remains almost constant 
and stable in different ZK-SNARK setups. This is due to the load of generating 
the proof being divided among the clients and executed in parallel. In contrast, the configuration 
without verification is consistently the fastest, as it does not require generating 
any proof. 
Although parallel execution over multiple clients limits epoch estimation time from increasing significantly with increased client number, the computational performance required to perform complex proofs remains a significant concern. 

 \subsubsection{Blockchain comparison}

Figure~\ref{fig:epochestimationtime} it states that the epoch estimation time of 
the blockchain remains nearly constant even as the number of clients increases, 
in contrast to zk-SNARK setups, where the epoch estimation time increases as the 
number of clients increases. However, the increase is more significant in circuits 
with more complexity, whereas the no verification
setup remains consistently faster because no proofs are required to be generated.
  
\subsection{Real epoch estimation time analysis}

\subsubsection{Scalability}

Figure~\ref{fig:realepochestimationtime} depicts real-time epoch estimation. The results depict that with increasing number of clients from 1 to 100, the real epoch execution time decreases.. This is due to the workload being divided across multiple clients, thereby the computation time per client is reduced. The results also indicate that the real-time epoch remains relatively high across all zk-snark setups. 
Larger circuits consistently result in longer epoch times, as clients require more time to compute and generate proofs. In contrast, the configuration without verification is always the fastest, as no proof needs to be generated.
Hence, parallel execution enhances performance as the workload is divided and handled in parallel, which leads to a reduction in real-epoch time .
These results demonstrate that real-time scalability in zk-SNARK configurations is enhanced by parallel execution, but remains limited by the costs associated with generating complex proofs. 


\subsubsection{Blockchain comparison}
  
Figure~\ref{fig:realepochestimationtime}  indicates that the real epoch estimation
time of blockchain stays almost constant as the number of client increases, compared 
to the zk-SNARK setups, where the time is decreases as the number of clients increases.
This difference is caused because blockchain only records updates without generating
proofs, compared to zk-SNARK setups that require each client to generate a proof for
each batch. However, the increase is more noticeable for more complex circuits, as
proofs of 1,000 variables require more computation than proofs of 500 or 700 
variables. Conversely, configuration without verification is still systematically
faster because no proofs need to be generated.


\subsection{Proof and verification time analysis}

\subsubsection{Scalability}
Figure~\ref{fig:proofandverificationtime} the experiment tests the proof generation
and verification time for ZK-SNARK with various numbers of variables ranging from 
(500, 700, 1000) from 1 to 100 clients. It is noticed that the proof generation and
verification time increases as the number of clients is increased for all ZK-SNARK
setups. The reason for this is that a higher number of clients need more time to
generate proof and verify it.
This observation reflects the reality that the number of clients and the complexity of the proofs both have an impact on performance. With an increase in the number of clients, there are more parallel computations, while the time needed for generating and verifying larger proofs still increases significantly. This underlines the difficulty of ensuring efficiency as the number of clients and the size of the circuit increase. 
\subsubsection{Blockchain comparison}

Figure~\ref{fig:proofandverificationtime}  indicates that the blockchain proving 
and verification time remains low and nearly constant as the number of clients
increases, compared to zk-SNARK configurations, which require more time to verify 
as the number of clients increases, especially with larger proof sizes. Since only
records are updated on the blockchain and no proof verification is performed. 

\section{Conclusion}\label{sec:conclusion} 

This research paper proposes a verifiable split learning architecture integrated with ZK-SNARK to ensure correctness and verifiability. The scalability of ZK-SNARK was analyzed under various circuit complexities and client numbers and then compared with the blockchain. The experimental results showed that zk-SNARKs offer strong guarantees that computations are performed correctly and honestly, but are slower. Blockchain, on the other hand, is faster and easier, but does not allow verification that computations have been performed and verified correctly.
Future work could focus on improving the computation circuit and investigating lightweight zero-knowledge proofs, and trade-offs by using gradient concealment and sparsification to reduce the computational load.

\bibliographystyle{IEEEtran}
\bibliography{references}


\end{document}